\documentclass[11pt, a4paper]{article}

\usepackage{booktabs}

\usepackage{PRIMEarxiv}
\usepackage{multicol}
\usepackage{listings}
\usepackage{fancyvrb}
\usepackage{spverbatim}

\usepackage[utf8]{inputenc} 
\usepackage[T1]{fontenc}    
\usepackage{hyperref}       
\usepackage{url}            
\usepackage{booktabs}       
\usepackage{amsfonts}       
\usepackage{nicefrac}       
\usepackage{microtype}      
\usepackage{float}
\usepackage{lipsum}
\usepackage{fancyhdr}       
\usepackage{graphicx}       
\usepackage[export]{adjustbox}
\usepackage[font={small,it}]{caption}
\usepackage{lipsum}
\usepackage{changepage}   
\graphicspath{media}     

\title{MANTA - Model Adapter Native generations that's Affordable}

\author{
  Ansh Chaurasia \\
  University of California, Berkeley \\
  \texttt{achaurasia@berkeley.edu} \\
}

\begin{document}
\maketitle

            \begin{figure} [htbp]
                \centering
                \includegraphics[width=0.5\linewidth]{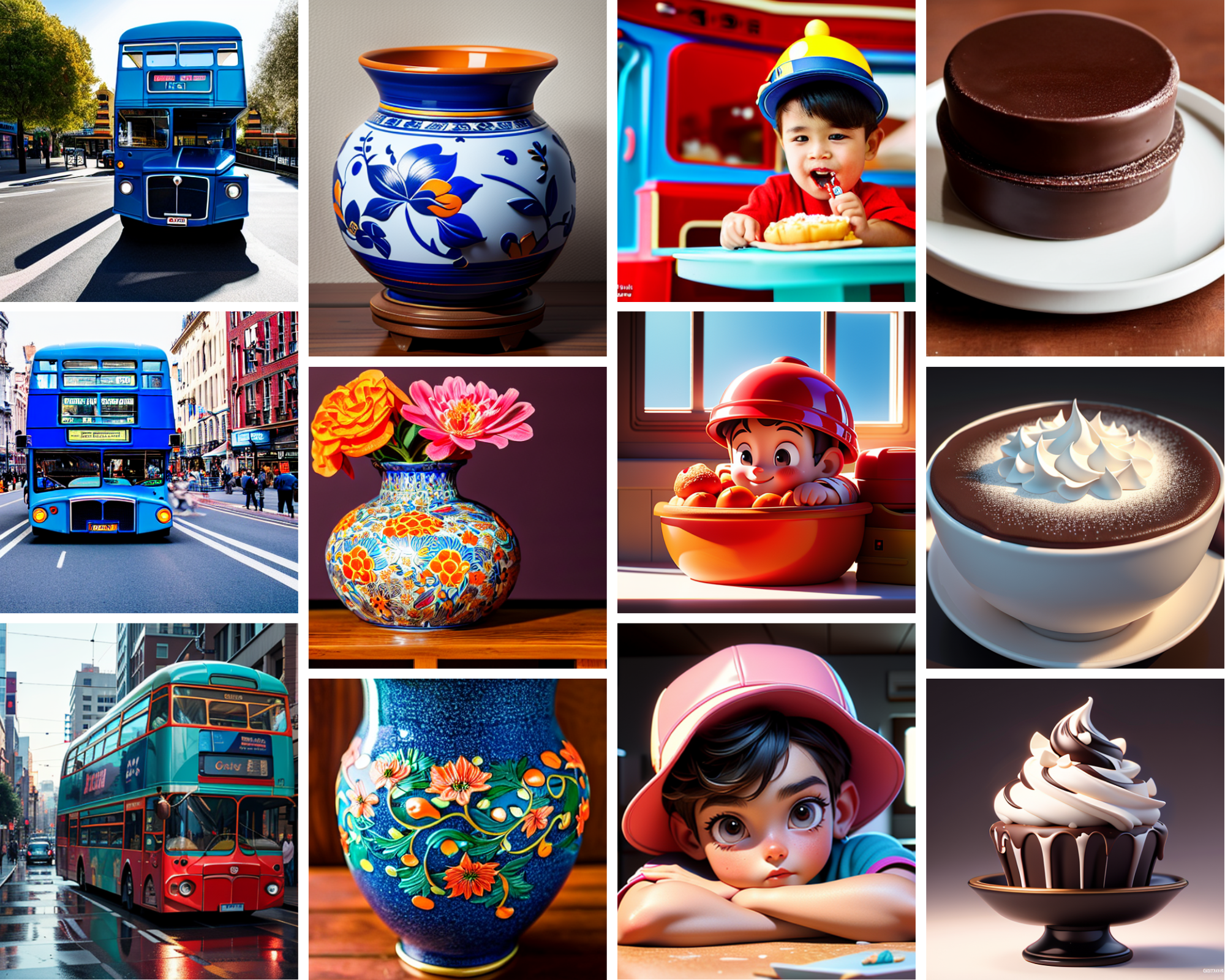}
                \label{fig:cover}
            \end{figure}
\begin{abstract}

The presiding model generation algorithms rely on simple, inflexible adapter selection to provide personalized results. We propose the \textit{model-adapter} composition problem as a generalized problem to past work factoring in practical hardware and affordability constraints, and introduce MANTA as a new approach to the problem. Experiments on COCO 2014 validation show MANTA to be superior in image task diversity and quality at the cost of a modest drop in alignment. Our system achieves a $94\%$ win rate in task diversity and a $80\%$ task quality win rate versus the best known system, and demonstrates strong potential for direct use in synthetic data generation  and the creative art domains .

\end{abstract}

\section{Introduction}

\begin{multicols}{2}
\subsection{Background}

Since the recent popularization of diffusion models by Ho et al \cite{ho2020denoisingdiffusionprobabilisticmodels}, significant work has been undertaken into creating AI models that can be easily harnessed by users in generating images for a specific, custom use case \cite{gal2022imageworthwordpersonalizing} \cite{ha2016hypernetworks}. Stable Diffusion provided the first contemporary example of a \textit{text-to-image}  latent diffusion model, i.e, a diffusion model that could be conditioned using embeddings from a text encoder, operate with memory-efficiently using a compressed latent space, and provide extremely high resolution and quality images. With the popularization of Stable Diffusion, users gained the capability of being able to finetune a base checkpoint Stable Diffusion model for additional customizability. 

However, the pretrain-then-finetune approach \cite{liu2021pretrainpromptpredictsystematic} was still computationally expensive, and broadly infeasible to the general public. The creation of parameter efficient finetuning methods such as Low Rank Adapation (LoRA) and Parameter Efficient Finetuning (PEFT) attempted to tackle this shortcoming, popularizing the  \textit{adapter} paradigm, where one could more easily create an adapter co-existing as an addendum to the model that would provide the necessary customization at a fraction of the computation. 

The current state of the art comprises of pairing image generation model checkpoints with additional adapters such as LoRAs to construct an image generation workflow. However, finding an appropriate combination of models and checkpoints continues to remain an open challenge, particularly useful for synthetic data generation, where additional data can be used to augment the data distribution \cite{bauer2024comprehensiveexplorationsyntheticdata}, or creative AI art \cite{tang2024exploringimpactaigeneratedimage}.

Checkpoint and adapter selection is predominantly done manually, leading to very little exploration in finding model-adapter combinations that would address a custom workflow \cite{xiao2024omnigenunifiedimagegeneration}. Oftentimes, users may attempt to pick from a series of existing popular models, experiment with the model to understand it's capability for an image concept, and then find adapters to enhance the quality of the generated images.

With the advent of large-language models, the retrieval augmented generation (RAG) paradigm has become extremely common for systems attempting to find the most relevant content related to some given input. This typically consists of a large language model, some input source documents, and a query. Given the query, the retrieval augmented generation system searches for the most relevant documents from the sources, appends it to its response, and then attempts to answer the question. 

\subsection{Core Research Problem}
Previous retrieval based systems (Stylus) \cite{luo2024stylus} defined their system as solving the \textit{adapter composition problem}. Given a prompt $P$ and a fixed model $C$ and a set of adapter  $A = \{ L_1, L_2, ... L_k\}$ how can one find a set of adapters $(\{ L'_1, L'_2 , ... L'_n\})$ that would systematically improve \textit{image diversity} while having model output generation $O$ retaining \textit{alignment} to $P$ . 

This line of research attempts to generalize this to the broader  \textit{model-adapter composition problem} by assuming there are additional choices to make for the model $C$ as well. Given a prompt $P$ , a set of models $C = \{ C_1, C_2, ... C_k \}$ and the adapter set $A = \{ L_1, L_2, ... L_k\}$ , where any $L_i$ is capable of providing additional fine tuning to any of the models in $C$, how do we come up with a system $S$ that effectively maps $P$ to a set of adapters \textit{and a model} $( C_i, \{ L'_1, L'_2 , ... L'_n\})$ such that the new combination provides additional output information. In the image domain, this "additional information" typically refers to output diversity or quality, commonly measured objectively through the Frechet Inception Distance (FID) or Inception Score (IS), while more emerging holistic, human-approximating methods use Vision Language Models such as GPT-4V.

Furthermore, this research paper extends the latter problem of model-adapter composition in a standard, lower-end consumer grade hardware setting. We attempt to extend the knowledge of previous work towards also selecting the most appropriate checkpoint in the image domain case. Additionally, we acknowledge that there are various \textbf{software and hardware budget constraints} \cite{silvano2024surveydeeplearninghardware}, and therefore additionally define the token budget $T$, representing the number of tokens sent to AI mechanisms for embedding or generation purposes \cite{wang2024reasoningtokeneconomiesbudgetaware}. Additionally, as a large portion of the image generation community relies on limited, consumer-grade hardware, we seek to conduct experiments and list out performance over hardware profiles that AI hobbyists and enthusiasts would find useful when considering accompanying hardware.

\subsection{Past Work}

Research within the diffusion model based image generation domain continues to move at a breakneck pace: customization of output through adapters has happened so far in two notable manners - (1) direct integration at the model level, or (2) retrieval based methods to directly apply adapters on top of checkpoints. To the best of our knowledge, we have not seen any works delving directly into checkpoint selection.

While model based methods have proven to create higher quality images, we note that these are extremely unfeasible at a large scale, due to direct augmentation of the model causing storage constraints; many of these papers also assume a well defined set of adapters, which is not a steadfast requirement in our case. 

We list relevant work addressing model customization for more custom output. 
\begin{figure}[H]
    \centering
    \includegraphics[width=1\linewidth]{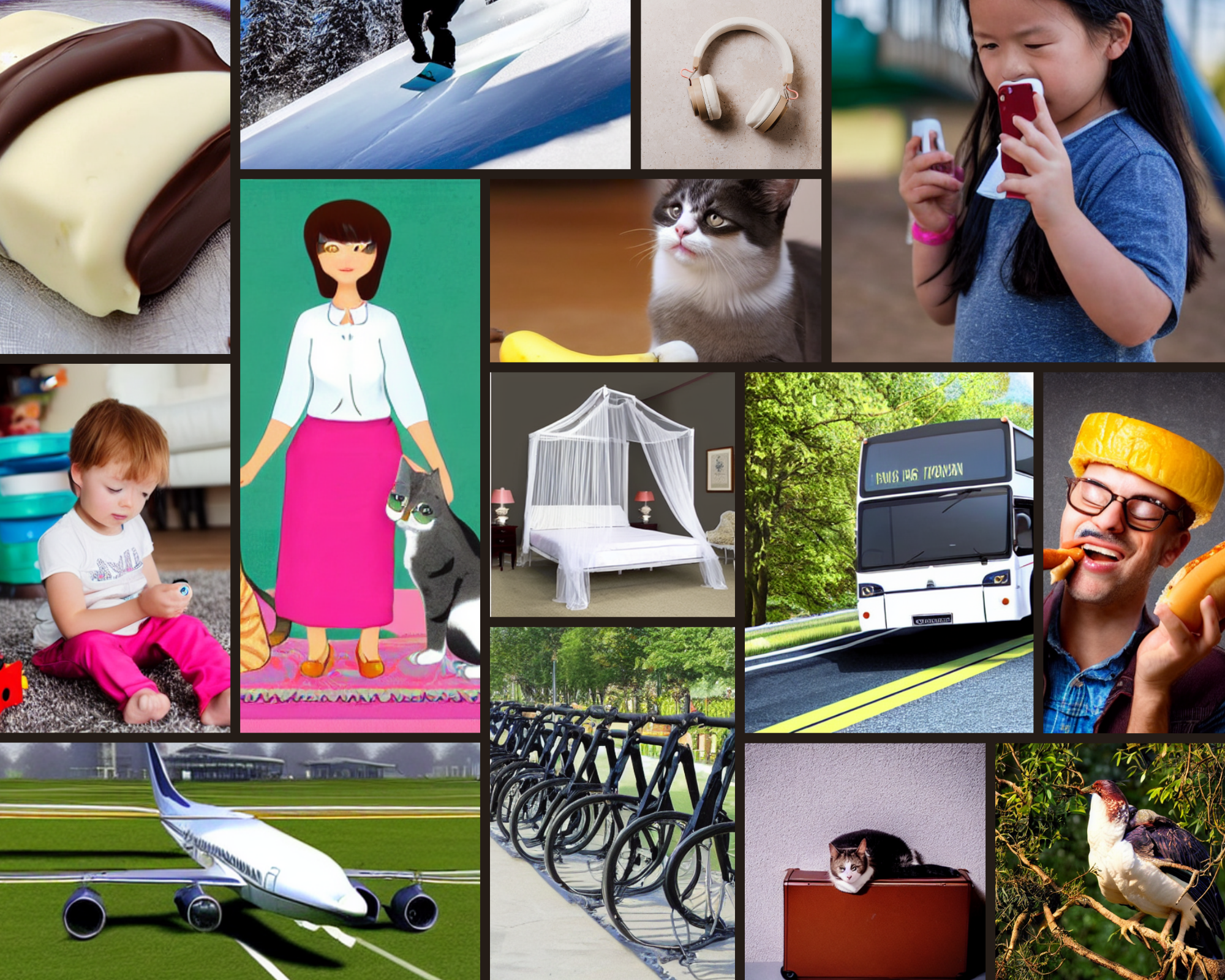}
    \caption{Examples of images generated via Stylus}
    \label{fig:past-examples}
\end{figure}
\subsubsection{Model Based Methods} 

Gu et. al \cite{gu2023mixofshowdecentralizedlowrankadaptation} created the Mix of Show algorithm that directly updates pretrained models at the weight level in order to resolve concept conflicts, as well as to bring extremely unrelated concepts together. While this enhanced cohesion does improve image quality, directly updating a pretrained model template's weights per LoRA combination or prompt would be unuseable at a large scale, and time consuming. In another direction, Choi et. al \cite{choi2024simpledropinloraconditioning} demonstrate improved adapter performance (and thus, customizability) at the architecture level, applying adapter weights to a specific portion - the attention layers - rather than the entire model. 

In another recent work, Yang et. al \cite{yang2024loracomposerleveraginglowrankadaptation} creates the LoRA Composer system, which promotes the \textit{training free} adapter composition approach that we similarly attempt to advance. Rather than altering the model weights, LoRA composer promotes inference time ideas such as latent re-initialization and constraining the latent space to effectively operate for a subsection. 

\subsubsection{Retrieval Based Methods}
The first retrieval based adapter selection approach known came from Luo et. al \cite{luo2024stylus}, who developed Stylus (referred to as Stylus or Stylus 1.0 throughout the paper). The system was created to resolve the Adapter Composition problem using retrieval augmented generation (RAG) with downstream re-ranking. This was done by fixing a base model and using a large language model (LLM) to compose adapters in the form of Low Rank Adaptations (LoRAs) based on found titles and description metadata that appear to be the most relevant \cite{setty2024improvingretrievalragbased} \cite{dong2024dontforgetconnectimproving}. As previously stated by the authors in their abstract - \textit{Stylus outlines a three-stage approach that first summarizes adapters with improved descriptions and embeddings, retrieves relevant adapters, and then further assembles adapters based on prompts' keywords by checking how well they fit the prompt.}\cite{luo2024stylus}

Some core contributions from their paper include (1) introducing an early framework for \textit{concept mapping}, and then (2) associating each adapter with the concept-mapped keywords that can reframe the problem into a retrieval augmented generation situation \ref{fig:past-examples}. 

\subsection{Retrieval Methods Limitations and Opportunities}

With Stylus, there were multiple open challenges revealed in the area of systematic model composition. We first discuss the limitations and failure modes seen from our evaluation of the Stylus system from the image domain perspective and then in the broader attempt at adapter composition.

\subsubsection{Lack of Task Diversity}
The Stylus system heavily relies on metadata such as descriptions, titles, and other textual metadata in its retrieval mechanism. While this does prove functional in practice, we find that this commonly leads to improper output generation and low alignment, a problem pervading image generation systems in general \cite{sadat2024cadsunleashingdiversitydiffusion} \cite{zameshina2023diversediffusionenhancingimage} \cite{lin2024diffusionmodelperceptualloss} \cite{Park_2024}.  

\begin{figure}[H]
    \centering
    \includegraphics[width=1\linewidth]{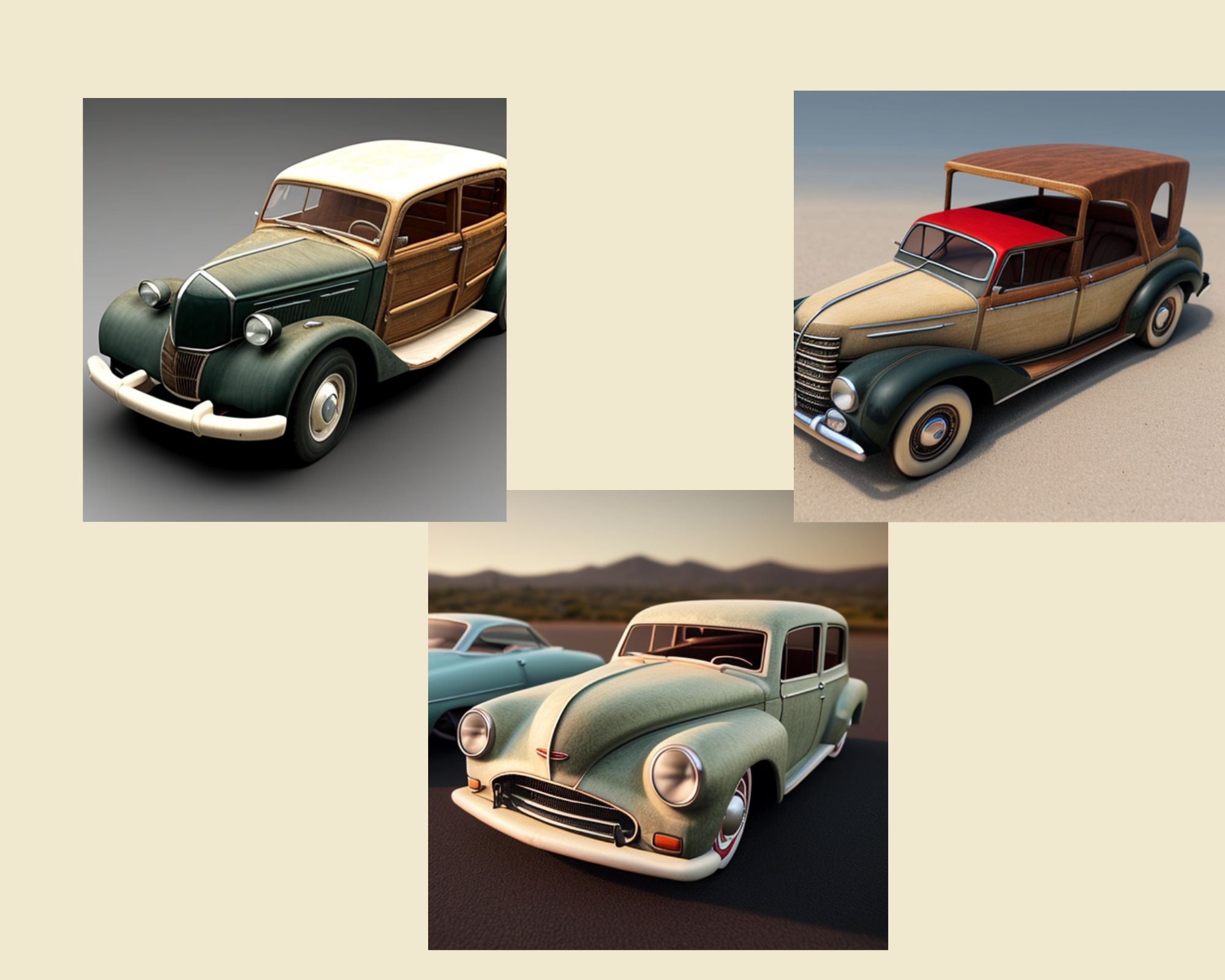}
    \caption{Example of a "low image diversity" generation, source Stylus. The majority of the cars synthetically generated look extremely similar and generic, and all have muted backgrounds. }
    \label{fig:low-diversity}
\end{figure}

Stylus fosters image diversity through pure randomness, randomly drawing permutations of LoRAs that seem be relevant within reason. While this method did foster diversity, there are visible limitations to the extent of which this diversity reaches, which stems from the lack of vetting adapters with previously tested examples of output \ref{fig:low-diversity}. 

.

We improve upon the diversity in our system, MANTA, inserting an additional step to find the most appropriate checkpoint to further image diversity, while minimizing token usage and RAM requirements by leveraging previous image prompts using the model as source documents.

\subsubsection{Low Alignment}
\begin{figure}[H]
\centering
    \includegraphics[width=0.5\linewidth]{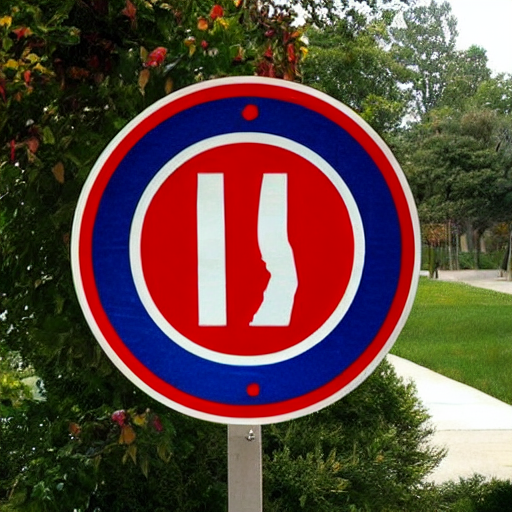}
    \caption{Example of a low alignment output from Stylus. Prompt: A stop sign that has the picture of George Bush in place of the letter O.}
    \label{fig:low-alignment}
\end{figure}

Similarly, there are examples where the previous image generation system created images with little consideration for how the concepts in the image seek to interact with one another \cite{wallace2023diffusionmodelalignmentusing}. This leads to images which may be seen as diverse, but in an unintentional and negative manner \ref{fig:low-alignment}. 

\subsection{Current Image Generation Workflow Challenges}

\subsubsection{Image Resolution}
In developing image generation models, a key component is requiring high image quality at large resolutions. In the context of AI generated images, typical requirements include images with sizes of at least 512 x 512, reduced blurriness and graininess, and a general lack of inconsistencies typically covered through a negative prompt - disfigured faces, malformed body parts (ex: a hand with six fingers), etc \cite{ha2024organicdiffuseddistinguishhuman} \cite{Lattimer_2023}.

\subsubsection{Alignment}
Image generation models facilitate achieving prompt-output alignment in image generation models for art through attention scaling \cite{newton2023aiartindustrialrevolution}. If the base prompt is insufficiently detailed, the model may struggle to incorporate all desired subjects effectively, resulting in incomplete or imbalanced images \cite{prabhudesai2024aligningtexttoimagediffusionmodels} . Determining the right combination of prompt elements is crucial to ensure comprehensive and coherent representation of the intended subjects. This involves fine-tuning the attention mechanisms within the model to adequately emphasize each aspect of the prompt. Addressing this issue is vital for producing high-quality, cohesive AI-generated art that faithfully reflects the user’s intentions.

\subsubsection{Image Diversity}
Consumers of generative models also typically look for control over image diversity, which refers to being able to create images with configurable amounts of variance \cite{naeem2020reliablefidelitydiversitymetrics} \cite{lin2024diffusionmodelperceptualloss}. 

During the start of the process, when users may be more focused on creating ideas, they typically seek to create a large number of images with higher variance to find a concept that they may seek to pursue. Typically further in the process, they then seek to curb variance to delve deeper and add further detail into an image previously selected during ideation.

\subsubsection{Consumer Friendliness}

AI art users have a consistent and well defined defined hardware profile that often is significantly different from research assumptions of availability of high computational resources\cite{stablediffusionbuildpc} . Rather, systems running AI art often contain GPUs with VRAM \cite{zhou2024surveyefficientinferencelarge} ranging between $8$ GB and $24$ GB, and RAM of about $96$ GB on the upper end. 

Additionally, demand in the industry for configurable AI art systems has been on the rise, where developers seek frameworks that can be easily customized to handle their organization's unique usecases. Hence, we attempt to design MANTA with \textit{complete model configurability in mind}, providing adapters to re-configure any LLM used within our work to an open source LLM or in-house LLM that can further be finetuned. Features such as large context lengths may be expensive for a commercial startups to sustain, hence we design our system with a focus on minimizing the tokens used \cite{pilz2024increasedcomputeefficiencydiffusion}.

\section{Related Works}

\subsection{Adapters}
Adapters efficiently fine-tune models on specific tasks with minimal parameter changes, reducing
computational and storage requirements while maintaining similar performance to full fine-tuning \cite{gal2022imageworthwordpersonalizing, ha2016hypernetworks, hu2021loralowrankadaptationlarge}. 

However, an integral part of the image generation process is finding an appropriate foundational model checkpoints \cite{touvron2023llamaopenefficientfoundation} to complement these adapters. Our research focuses on locating better checkpoints and obtaining Low-Rank adapters (LoRA) that most appropriately align with the prompt, while maintaining the popular approach within existing open-source communities \cite{civitai, peft, yao2023treethoughtsdeliberateproblem}.

Adapter composition has emerged as a crucial mechanism for enhancing the capabilities of foundational models across various applications \cite{li2024playgroundv25insightsenhancing, podell2023sdxlimprovinglatentdiffusion, touvron2023llama2openfoundation} . In the image domain, combining LoRAs effectively enhances different tasks—concepts, characters, poses, actions, and styles—together, yielding images of high fidelity that closely align with user specifications \cite{liu2023unsupervisedcompositionalconceptsdiscovery, zhong2024multiloracompositionimagegeneration}. Our approach advances this further by actively segmenting user prompts into distinct tasks and merging the appropriate adapters for each task.

\subsection{Retrieval}
Retrieval-based methods, such as \textit{retrieval-augmented generation} (RAG), significantly improve model responses by adding semantically similar texts from a vast external database \cite{lewis2021retrievalaugmentedgenerationknowledgeintensivenlp}. These methods convert text to vector embeddings using text encoders, which are then ranked against a user prompt based on similarity metrics. Similarly, MANTA draws inspiration from RAG to encode adapters as vector embeddings. A core limitation to RAG is limited precision, retrieving distracting irrelevant documents. This leads to a ”needle- in-the-haystack” problem \cite{li2024retrievalaugmentedgenerationlongcontext}, where more relevant documents are buried further down the list. We leverage recent advances in RAG, namely triplet-loss based searching techniques, to effectively combat this.

\subsection{MANTA Overview}

Adapter selection presents distinct challenges compared to existing methods for retrieving text-based documents, as outlined in Section 2. First, computing embeddings for adapters is a novel task, made more difficult without access to training datasets. Furthermore, in the context of image generation, user prompts often specify multiple highly fine-grained tasks. This challenge extends beyond retrieving relevant adapters relative to the entire user prompt, but also matching them with specific tasks within the prompt. Finally, composing multiple adapters can degrade image quality and inject foreign biases into the model. Our retrieval mechanism, using a novel similarity computation outlined below, addresses the challenges above.

\end{multicols}
\clearpage
\section{Our Method}

\begin{figure} [H]
    \centering
    \includegraphics[width=0.75\linewidth]{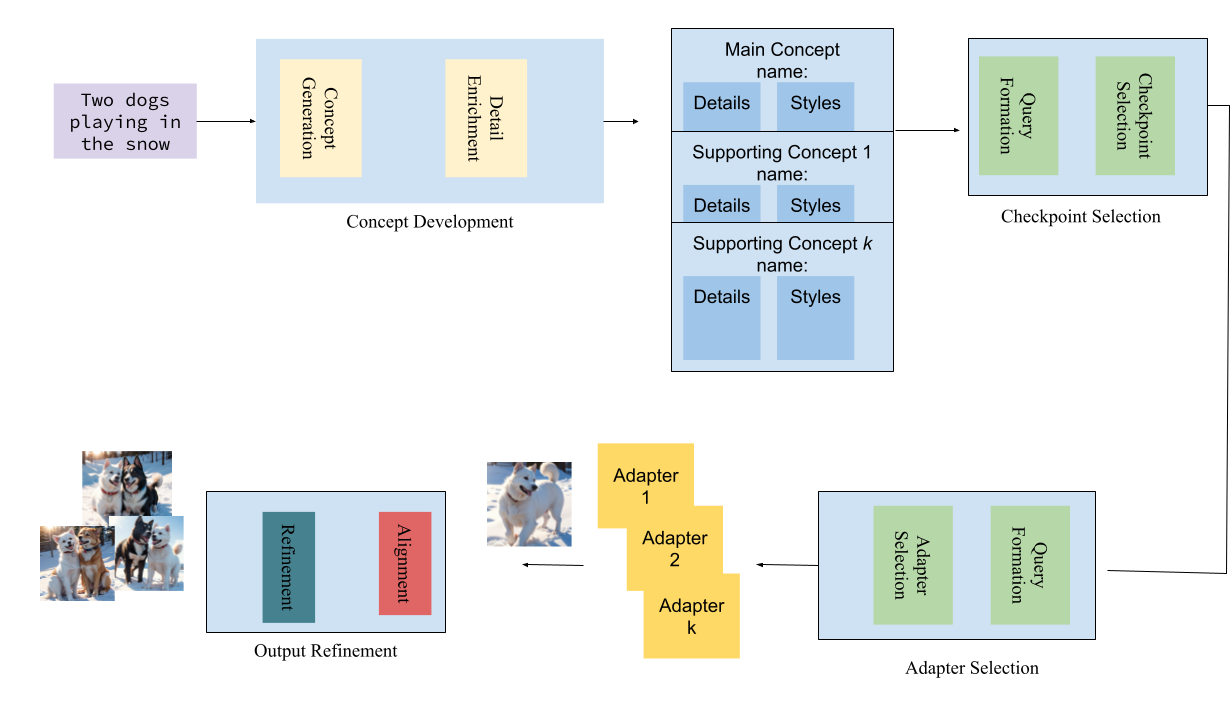}
    \caption{\textbf{MANTA} algorithm. The system consists of four stages - \textbf{concept development}, \textbf{checkpoint selection}, \textbf{adapter selection}, and \textbf{ refinement}. The output refinement procedure simply acts as a pass through for the time being, but serves as a location to insert alignment mechanisms.}
    \label{fig:algo-architecture}
\end{figure}

The core MANTA system consists of 4 major processes - \textbf{Structured Concept Development}, \textbf{Detail Enrichment}, \textbf{Strategic Adapter Selection}, and \textbf{Output Refinement} \ref{fig:algo-architecture}.

\begin{multicols}{2}
In the following sections, we will further explain the processes within each of these steps.

\subsection{Structured Concept Development}

In this process, we leverage LLMs to analyze a prompt, and segment the prompt into a concept with three attributes - a name, some details describing the concept, and the styles to generate this concept with. For example, if you are attempting to generate an image with the input prompt 'alien', then: 

\begin{itemize}
    \item The Concept Name would be 'alien'
    \item Details could include: full body, alien creature, three heads, glowing torso, ...
    \item Styles could include: anime style, futuristic, ...
\end{itemize}

Additionally, we are able to create priorities between the subjects within the prompt by classifying each concept we identify as a \textit{main subject}, or a \textit{supporting subject}. Main subjects are given extreme priority throughout the process, and are used to strategically determine core components of the workflow, such as the best checkpoint, and optimal adapters.

An example of what this process would take would be an example prompt - 'i.e', 'a techno samurai warrior walking his cyberpunk dog', and return a dictionary mapping the \textit{main} and \textit{supporting} subjects.

\begin{spverbatim}
{
    'main': {
        'name': 'techno samurai warrior', 
        'styles': [], 
        'details': []
    }, 
    'support': [
        {
            'name': 'cyberpunk dog', 
            'styles': [], 
            'details': []
        }
    ], 
    'image': {'styles': [], 'details': []}
}

\end{spverbatim}

In the example above, the techno samurai warrior acts as the main concept, with the cyberpunk dog acting as a support. 

\subsection{Benefits of Structured Concept Development}

By using the Structured Concept Development framework to analyze prompts, we sought to provide the following benefits. 

\textbf{Systematic Variance Insertion}: As illustrated in the next section with \textit{Detail Enhancement}, structured concept development provides an avenue to systematically insert controlled amounts of variance, either via checkpoint or adapter, and impact image diversity. 

For example, while the artist may have a vague idea of what he wants to pursue in his mind, he would need to move forward with picture that is extremely vivid or well defined to create a high quality output. In practice, this means that the user may specify a short, vaguer prompt to the system and then iteratively obtain a clearer idea of what they want, and thus send lengthier, more comprehensive prompts.

Typically, that would require the user to feed a well defined concept, which can be achieved by using LLMs to additionally increase the details for the concept within the prompt. 

\textbf{Easy Interpretability for LLMs:} We have found that using a loosely structured approach to analyzing the prompt may be more effective than a completely unstructured approach (i.e, extracting key words or topics out of the input prompt) to ensure the relative significance of the concept is preserved.

From structured concept development analysis of prompts, concepts can systematically developed by LLMs by simply asking these LLMs to 'add similar and relevant details' to the list of existing details. 
 
\subsection{Detail Enhancement}

\begin{figure}[H]
    \includegraphics[width=1\linewidth]{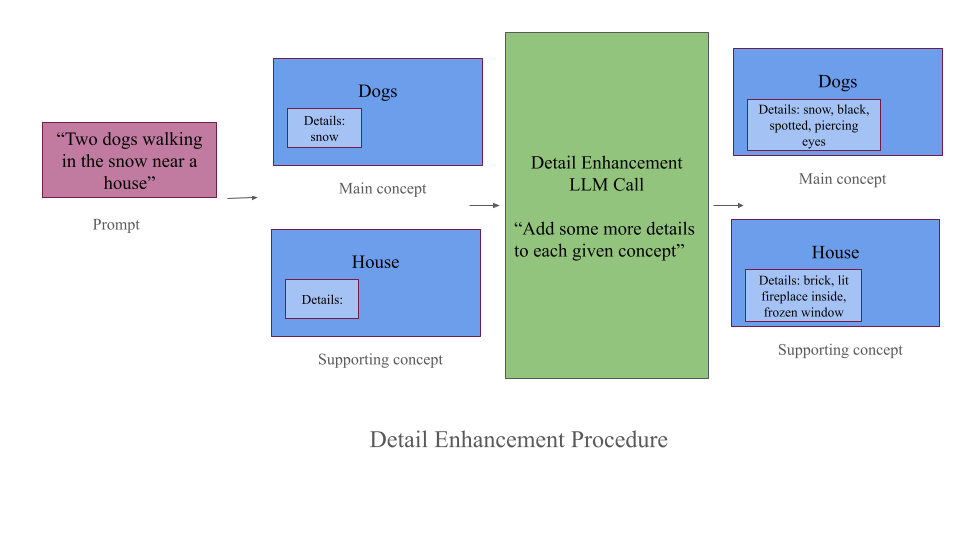}
    \caption{Overview of the detail enhancement process. The prompt is analyzed into a main concept and a set of supporting concepts, and then each concept is individually processed through the LLM to come up with more details.}
    \label{fig:detail-enhancement}
\end{figure}

We seek to design our system in a way that users can quickly ideate towards a "first concept" image, and then further refine. One of the core challenges in AI image generation is coming up with a first image "idea", which may further by refined upon. Generally in these cases, we see users using vaguer prompts, hoping to tap into the model's latent creativity to supplement them with ideas. 

However, this often leads to extremely vague images if the model isn't properly fine tuned for that specific idea. To address this, we enable the user to specify extremely vague prompts, and use LLMs to generate reasonable additional details based on the concept. The sample prompt is referenced here \ref{prompt:detail-enhancement}
.
Some examples of detail enhancement output for the prompt previously mentioned is shown below, for the techno samurai warrior concept.

\begin{spverbatim}
[
    'sleek metallic armor', 'glowing neon blue circuits', 'retractable energy katana', 'cybernetic enhancements', 
    'black visor helmet', 'steel-toed combat boots', 'metallic plate gauntlets', 
    'reinforced synthetic leather waist armor', 
    ...,
]
\end{spverbatim}

\subsection{Checkpoint / Adapter Retrieval}

\begin{figure}[H]
    \centering
    \includegraphics[width=1\linewidth]{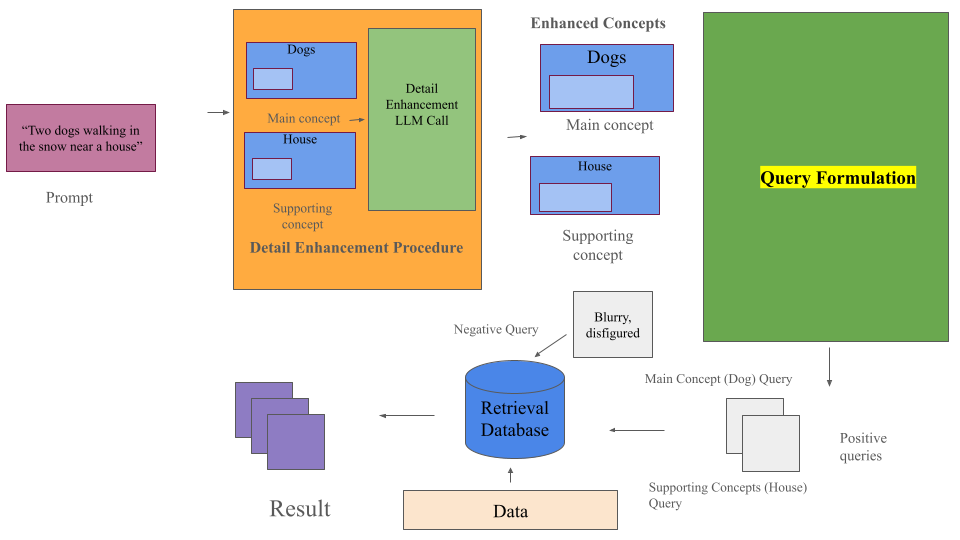}
    \caption{Overview of the retrieval process for some retrieval data - checkpoints and adapter information in our case. After the detail enhancement process previously listed, concepts are formulated as multiple queries and embedded (embedding now shown for brevity). Alongside a negative query, triplet loss is computed to find the most relevant adapters. }
    \label{fig:retrieval}
\end{figure}

The checkpoint and adapter selection mechanisms are identical, so we discuss the checkpoint selection mechanism to illustrate how the overall retrieval process functions. See Figure \ref{fig:retrieval} for a diagram  on the retrieval process.

Within the checkpoint selection phase, we attempt to locate a set of relevant checkpoints that are sufficiently relevant to the core concepts inserted into the prompt, that is if there exist a set checkpoints $C$ that have a relevancy score beyond the set relevancy threshold, $\omega_c$. 

Checkpoint Selection is divided into two core steps, \textit{document generation}, and \textit{checkpoint retrieval}. We leverage Qdrant \cite{pan2023surveyvectordatabasemanagement}, a state of the art vector database due to its strong interoperability with local and cloud deployments as the underlying vector database, which we insert to during the document generation, and query during retrieval.

\subsubsection{Checkpoint Document Generation}
In this section, we discuss the process through which source documents for checkpoints are generated. 

\textbf{Past Challenges}: The Stylus system attempted to create documents composed of titles, descriptions, and other user provided metadata as the primary point of search for source document construction. 

However, we find that we are able to achieve comparable or better performance by querying on the basis of image prompts. That is, we define the checkpoint retrieval problem as finding \textit{prompts that articulate our main concept} in similar fashion / context to the user inputted prompt.

\subsubsection{Checkpoint Document Retrieval}

To motivate our arrival at solely using image prompts for retrieval, we start back at the previous attempts of retrieval mechanism, starting from the predecessing system: 

\textbf{Iteration 1 - User Supplied Metadata}: Stylus provided a complete list of user metadata. However, this was extremely token-expensive computation. Descriptions would easily consume over $500$ words, and even with limiting the characters, often didn't reflect the checkpoints that were the most performant. 

An common example of this popped up when trying to generate animal based images. The concept would often include 'dog', 'cat', and other creatures semantically related to animal. However, due to the composition of the platform we tested with - CivitAI - we found that this would often pull in checkpoints dedicated to the topic of 'furries' (humanoid-animal hybrids), which would then hijack the output image generation. 

\textbf{Iteration 2 - Hybrid Keyword + Embedding Search}: To improve the results, we attempted to perform a coarse substring match filter on core concepts, followed up with an embedding based search to find the most relevant. This ran into the challenge of not being a general solution - while this was effective for general prompts involving 'cats' and 'dogs', extremely specific concepts like 'white limousine' had a low likelihood of being in the filter.

An attempted fix was to use a generalization prompt, to repeatedly generalize the concept into a more abstract version of the concept, i.e 'car' to 'vehicle', however this failed because of LLM unpredictibility - it often outputted words such as 'transportation' that were more esoteric. 

\textbf{Iteration 3 - Multi embedding Search on Prompts}: To improve performance while minimizing tokens processed, we performed a multi embedding search, using the concept mapping to split the image into smaller, isolated queries that are then fed through a triplet loss-like function, containing positive and negative queries.

\[ context = \sum min(s(v^i_+) - s(v^i_-), 0.0)\]

Where $v^i_+$ and $v^i_-$ are positive and negative examples provided by us. In our experiments, we provided a standard negative embedding, combined with a positive embeddings from the concept map defined earlier. 

Specifically, for our checkpoint selection, we split our concept map into queries of the following format:

\begin{spverbatim}
    QUERY FOR CONCEPT = concept_name + " " + concept_details + " " + concept_styles
\end{spverbatim}
This query is formulated for the main concept, and the support concepts are added as well to ensure a generally relevant checkpoint and LoRA. Additional query strategies and their corresponding results have been discussed within our Ablations section.

\section{Results}

\subsection{Experimental Setup}

Listed below are the key components used for experimental setup, followed by the experimental procedure used for each individual setup. 

\textbf{Adapter solicitation.} Adapters were solicited from civitai.com via API, for which previous authors for the Stylus page created a dataset known as StylusDocs, containing 75K adapters for various Stable Diffusion models. These adapters were stored as embeddings alongside a JSON file containing textual content, which used to populate the Qdrant vector database collections for information on LoRA adapters, as well as checkpoint adapters.

These vector databases were then quantized via INT8 scalar quantization in order to fit across RAM and 1 Tesla T4 GPU. 

\textbf{Base Image Generation Models.} The predecessing paper opted to use Stable Diffusion 1.5 models as their default, and for the sake of thoroughness and providing results relevant to today, we provide results tested with both Stable Diffusion 1.5, and SDXL, a newer standard that has begun to take over the AI art community. 

\textbf{Base Large Language Models (LLM)s.} For the sake of thoroughness, we primarily relied on OpenAI's gpt-4o model for all LLM generations. However, we also leverage a tool known as liteLLM, which enables users to drop in any model they seek in place of OpenAI, such as Google's Gemini API \cite{geminiteam2024geminifamilyhighlycapable}, or a local Ollama deployment. 

\textbf{Generation Setup.} In order to access Stable Diffusion models and create AI images in a consumer grade setting, the Automatic1111 stable diffusion web user interface was used as an API service. Two core operations were used from the API, which have both been listed below:

\begin{itemize}
    \item \textbf{Generation from Prompt:} This method was used by Stylus to be able to create it's images. Internally, given a prompt and negative prompt, the method leveraged Automatic1111's text-to-image generation interface to create images to share with the user. 
    \item \textbf{Generation from Image:} This method was used by Stylus to refine existing images that used the \textit{Generation from Prompt} method previously to refine and improve images. The method leveraged Automatic1111's img-to-img generation interface to refine images previously generated via text in the system.
\end{itemize}

\textbf{Hardware}
To simulate the experience of the average AI art user, we sought to replicate their environment by leveraging similar hardware. To that extent, we altered our hardware setup to use the following: 

\begin{itemize}
    \item \textbf{GPUs:} Under the belief that consumers tend to use lower end GPUs ranging from 8 GB to 24 GB on average, we opted for a single 16 GB VRAM Tesla T4 to conduct our experiments. 
    \item \textbf{CPUs: } 4 CPUs were used in the testing process
    \item \textbf{RAM: } In order to simulate consumer low-budget conditions, consumer grade resources such as lower $16$ GB RAM.
\end{itemize}

Due to these hardware constraints, only a single instance of the Automatic1111 API was able to be run at one time, in comparison to standard research environment hardware such as the A100, where 4-5 APIs with simultaneous models being provisioned.

\subsection{Experimental Procedures}

\subsubsection{Automated Evaluation}
To simulate a standardized version of human preference, we continue to use the automated evaluation mechanism leveraging vision language models to judge groups of images.

Specifically, the automated evaluation helps us obtain the primary basis of comparison for our understanding of the system's capabilities in \textit{image quality, image diversity, and alignment}. 

We ran automated evaluation on a group of $500$ generation run on the COCO dataset, comparing the results between images generated by the original Stylus system, MANTA, and a normal, unaugmented-by-LoRAs Stable Diffusion model.

Below contain the model preference comparisons of MANTA versus the original Stylus and SD. In each evaluation, the model was fed a criterion for the three categories (diversity, image quality, and alignment), and then provided two sets of images - one from MANTA, and another from the second image generation system being compared. 

The vision language model was then asked to rate each of the images for that category, and based on the category, come up with a preference for either images contained within the MANTA group, or the basis of comparison. In the table \ref{tab:main-auto-eval}, the total percentage of the times Stylus won has been recorded.

\begin{table} [H]

\caption{Automated Evaluation MANTA Win Percentage}
\label{tab:main-auto-eval}
    \begin{tabular}{|p{1.4cm}|c|c|c|} \hline 
         &  Diversity&  Image Quality& Alignment\\ \hline 
         Stylus&  0.94&  0.8& 0.43\\ \hline 
         Base Stable Diffusion 1.5&  0.85&  0.75& 0.51\\ \hline
    \end{tabular}

\end{table}

\subsubsection{Human Evaluation}
Human evaluation was conducted over the same sample of $500$ runs on the COCO dataset prompts. $100$ of these outputs were selected for evaluation, and evaluated by $4$ separate testers. 

Testers were shown a choice of a batch of images created using the original Stylus algorithm, the MANTA algorithm, and the respective base model. Human testers were requested to focus on image diversity and quality. In other words, subjects were asked to come up with a response to the questions below: 

\begin{itemize}
    \item Which set of images appears to be more diverse (varying in content, style, theme)? 
    \item Of all the images shown, which image is the highest quality?
\end{itemize}

\begin{figure} [H]
    \centering
    \includegraphics[width=1\linewidth]{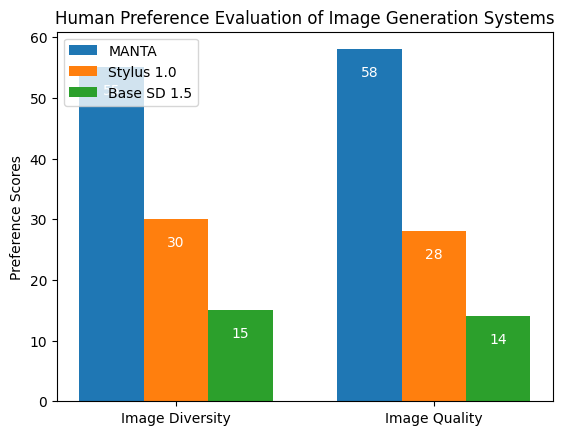}
    \caption{Results of Human Preference, ranked for three methods - MANTA, Stylus, and base Stable Diffusion}
    \label{fig:human-prefs}
\end{figure}

Based on these results, a \ref{fig:human-prefs} has been compiled of human image preference on image diversity on the $100$ COCO samples, as well human image quality preference. 

\begin{figure} [H]
    \centering
    \includegraphics[width=0.75\linewidth]{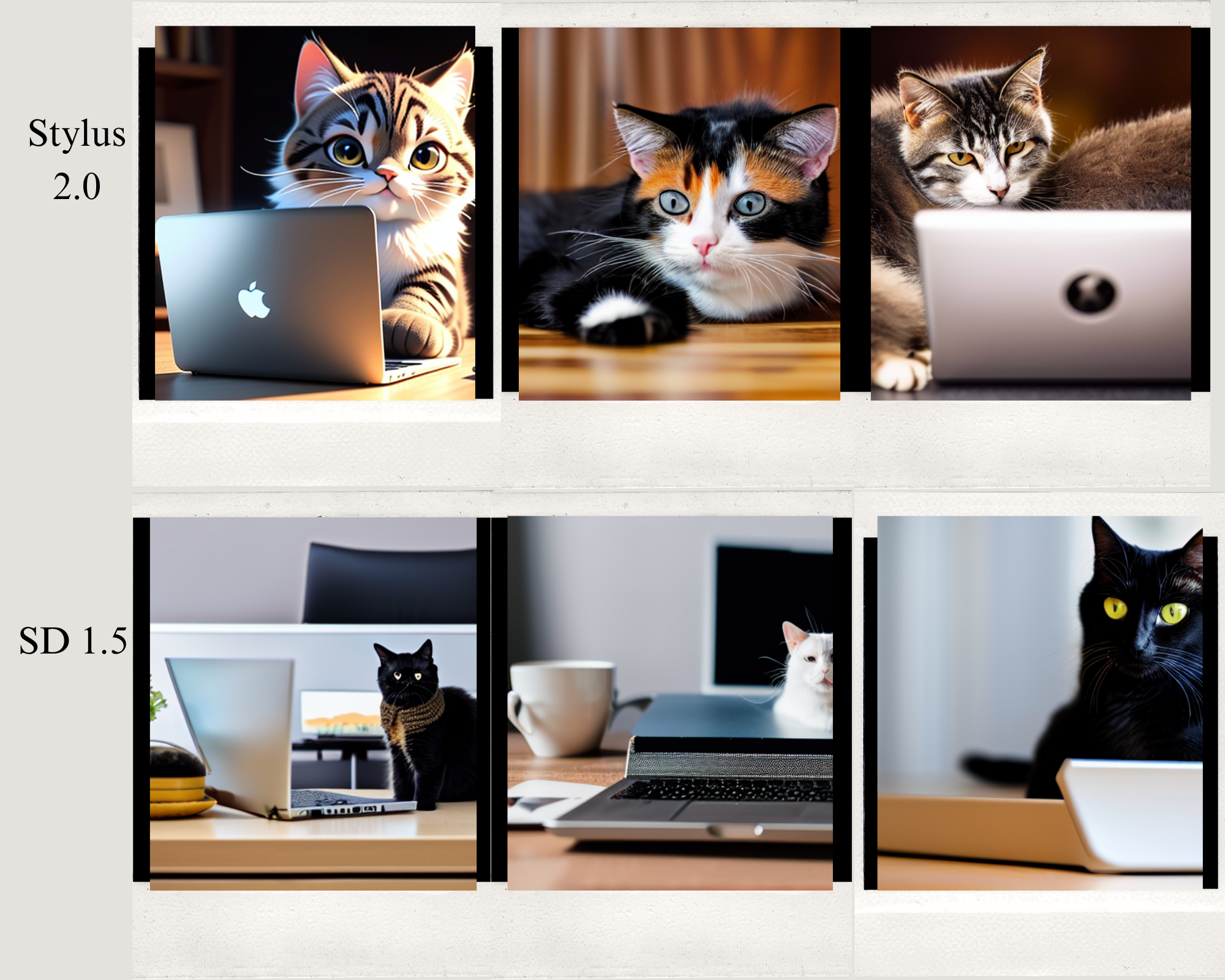}
    \caption{MANTA and base Stable Diffusion (SD) 1.5 outputs for the prompt \textit{A cat on top of a closed laptop on a desk}. There is clear variation in MANTA output, as opposed to extremely similar images from SD 1.5.}
    \label{fig:comparison-normal}
\end{figure}

Our human evaluation demonstrates a human preference percentage of $55\%$ in image diversity, and $58\%$ in image quality respectively for our MANTA algorithm, followed by a $30\%$ image diversity and $28\%$ win rate for the Stylus algorithm \ref{fig:comparison-normal}.

\begin{figure}[H]
    \centering
    \includegraphics[width=1\linewidth]{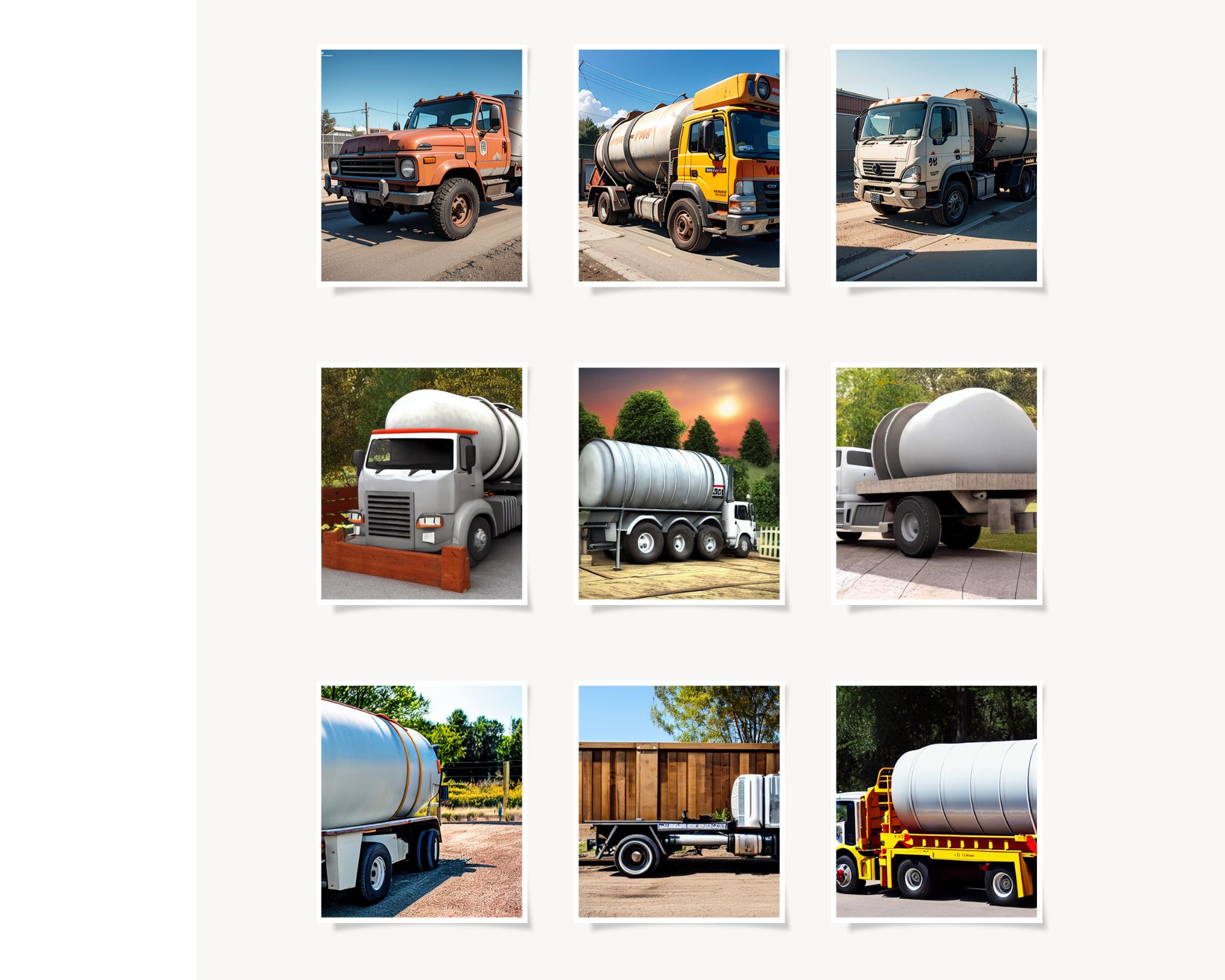}
    \caption{MANTA, Stylus, and normal Stable Diffusion outputs for the prompt: A cement truck sitting next to a fence. MANTA clearly demonstrates diversity across vehicle construction, color, and background situation while maintaining coherent relevance with other concepts.}
    \label{fig:manta-diversity}
\end{figure}

Human Evaluators observed that the MANTA algorithm regularly provided more prompt-specific diversity. For example, in the figure above, there is a clear variation in style between the three images created by MANTA - the first appears to have a fictional look, the second scene solely featuring the cat in a splayed pose, and the third, with the cat peering into the computer. In the example below \ref{fig:manta-diversity}, we see similar diversity versus Stylus and base Stable Diffusion. 
 
Conversely, we can see clear similarities between the images from default Stable Diffusion - the cat always appears to look off into the distance, a complete computer is never shone, and the background always appears to be some form of white rather than scenery.

On the other hand, alignment results seemed hard to confirm. The challenges we faced with alignment typically occured on complex prompts, where all 3 algorithms typically provided outputs that weren't very relevant to the prompt. Attached is a good example picked from the evaluation run, showing a better observance of the MANTA algorithm (following up from \ref{fig:low-alignment}). 

\begin{figure}[H]
    \centering
    \includegraphics[width=1\linewidth]{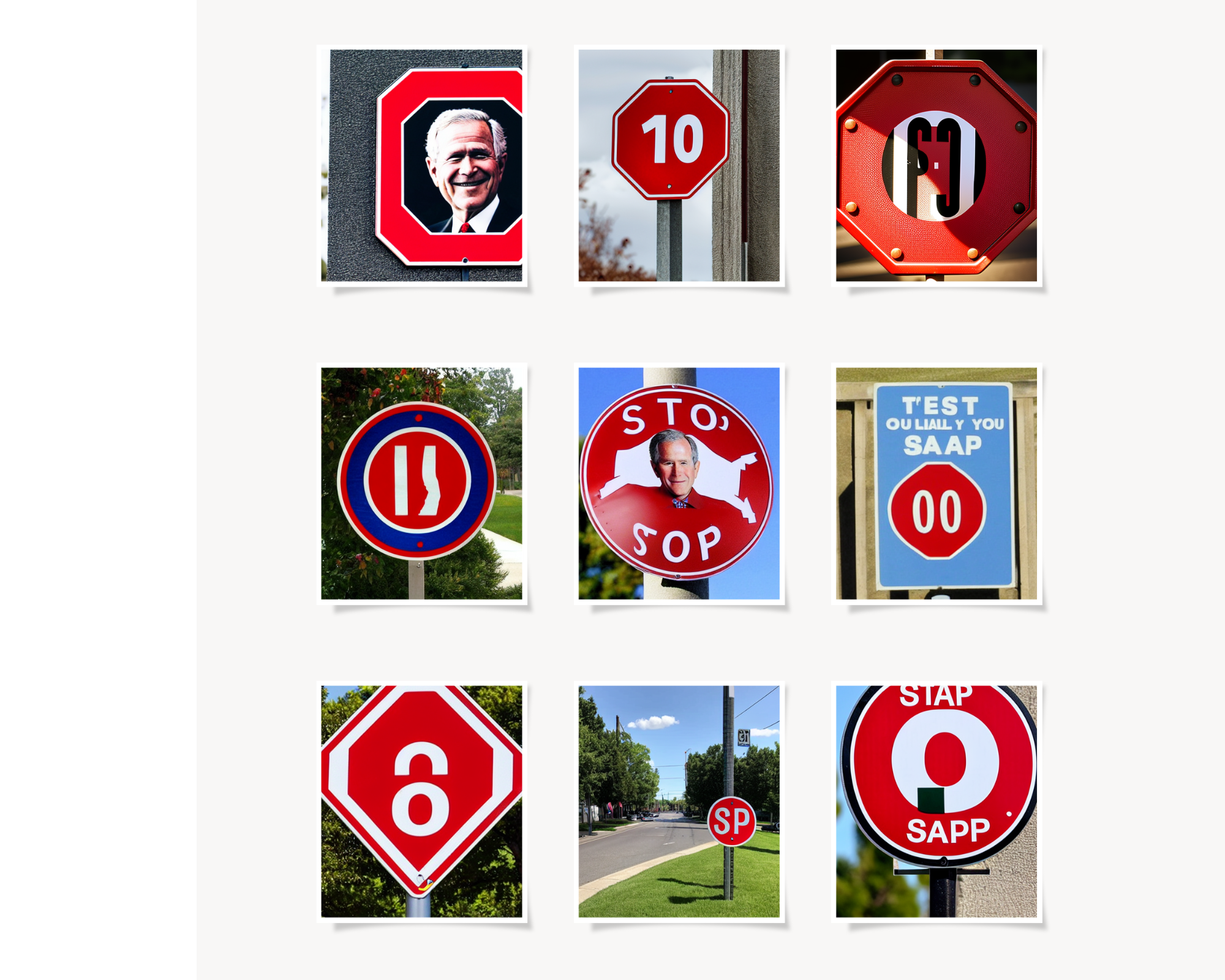}
    \caption{Example of an alignment output by the three algorithms - MANTA, Stylus, and base Stable Diffusion. Prompt: A stop sign that has the picture of George Bush in place of the letter O. As demonstrated, alignment is relatively low across all three images - none of the three are able to replace the letter 'O' with George Bush, but MANTA and Stylus do partially approach the prompt on 1 out of three images, with Stylus deviating further on that sample (inclusion of a rendition of the US). }
    \label{fig:alignment-challenges}
\end{figure}

\subsubsection{Token Count Comparison}
As commercial models were used in the system, approximate token counts per run are shown in order to help users estimate usage costs. While LLM generations are becoming significantly cheaper, the Stylus image generation systems still require large investments of LLM tokens for high quality output; therefore, we seek to provide an overview of the avenues of improvement MANTA seeks to provide in this area.

In order to estimate LLM token count, we leveraged OpenAI's tiktoken package. We tracked token usage by maintaining a running counter, which would track the number of tokens after every call to an LLM with a prompt, or with a request to embed textual information.

\begin{figure}[H]
    \centering
    \includegraphics[width=0.5\linewidth]{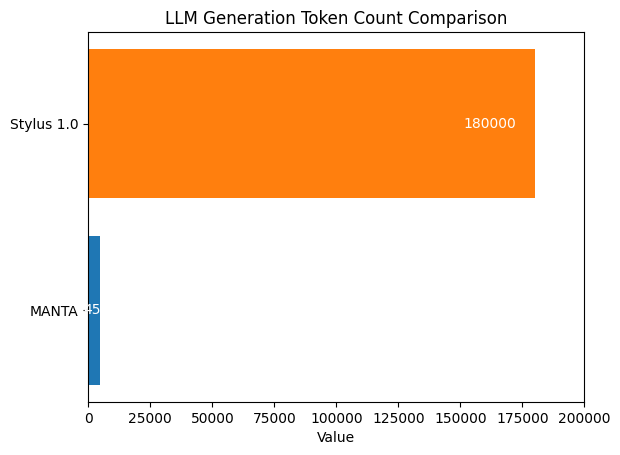}
    \caption{Graph of token count comparisons in LLM usage per image generated between Stylus and MANTA.}
    \label{fig:token-counts}
\end{figure}

Shown is a direct comparison of token counts from run using the previous Stylus iteration, versus the current run \ref{fig:token-counts}. 

As demonstrated by the graph, MANTA's average token count (to 3 significant figures) of $4500$ tokens provides approximately a $40$ times decrease from the original algorithm.    

We believe this significant decrease stems from two separate places - pre-processing, and concept generation. The past iteration depended on passing extremely large quantities of user metadata during the pre-processing to embedding models, resulting in extremely large token usage.

We replace user metadata about the checkpoints and adapters with example prompts users have previously used in the past with the weights, and attempt to query for weights which use concepts similarly in the prompt.

Additionally, we remove the reliance Stylus previously maintained on using a \textit{concept tagging} system, where each adapter would be associated with a series of tags, and then queried against. In contrast, the concept mapping framework is used once on the prompt to find the core parts, and then those core parts are queried to find similar prompts.

\subsection{Ablation Studies}

\subsubsection{Concept Enhancement}
For the purposes of a more direct comparison between MANTA and Stylus performance, we've included an ablation detailing the comparison between MANTA. In this ablation, both image generation models were fed the exact enhancement prompts from COCO 2014 \cite{lin2015microsoftcococommonobjects}, and then evaluated using automated GPT-4 \cite{openai2024gpt4technicalreport} evaluation. 

We obtained the following results, when testing generations with a sample size of $N=15$, replicated using the Stylus Docs benchmark. For the purposes of exploration, we explored the various modes the previous Stylus system had, and their performances. Note that we weren't able to replicate the  're-ranker' mechanism due to repeated internal errors, and were forced to switch it off:

\begin{table} [H]
    \centering
\caption{No-Concept Evaluation MANTA Win Percentage vs. Stylus}
\label{tab:no-concept-eval}
    \begin{tabular}{|p{1cm}|c|c|c|} \hline 
         &  Diversity&  Image Quality& Alignment\\ \hline 
         Stylus - Rank&  0.77&  0.94& 0.43\\\hline
 Stylus Random& 0.92& 0.89&0.31\\\hline
    \end{tabular}

\end{table}

The table \ref{tab:no-concept-eval} demonstrates the significant but non-encompassing impact of prompt enhancement via the loose concept framework. While there was a notable decrease in the diversity  ($-17\%$) and minor increase quality ($+4\%$ ) win rate against Stylus Rank, we  still see an remarkable win rate against Stylus, suggesting concept enhancement provided valuable improvement in diversity, but the system could still provide reliable image diversity in not used. We postulate that the gap comes primarily from checkpoint inductive bias - which we further explore in the next ablation study. By repeatedly using the same one model, we believe that the original Stylus system suffers from a significantly higher model-based bias, in comparison to a system that has the flexibility to alternatively consider from a sampled set of closely relevant checkpoints.

\subsubsection{Base Checkpoint Variation}

\begin{figure}[H]
    \centering
    \includegraphics[width=1\linewidth]{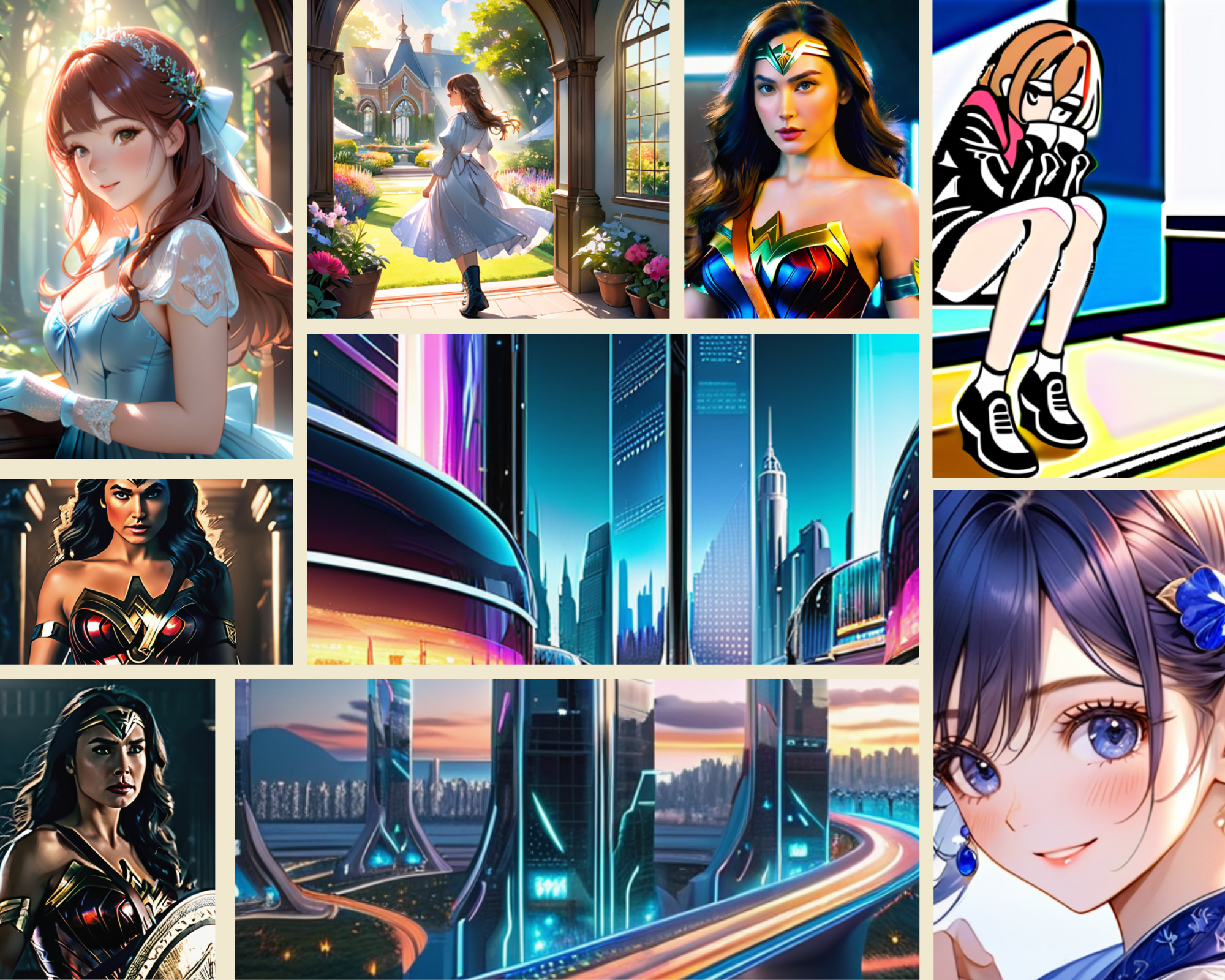}
    \caption{Images generated by Stylus with SDXL 1.0 as the checkpoint architecture.}
    \label{fig:enter-label}
\end{figure}

We attempted to obtain images from a second, larger AI model to understand their significance on MANTA output. Apart from the expected VRAM increase, we found multiple mounting challenges, such as the lack of checkpoint diversity and number of adapters for said checkpoints. 

That being said, there were sufficient checkpoint-adapter pairings within the anime, heroes, and landscape area to test out the first version of the state of the art SDXL image generation model. The images were better, but the quality jump versus standard SD 1.5, which has a more robust ecosystem, was still minimal. 

We hoped to test against the latest FLUX-series models released, however, our experimental setup (the Automatic 1111 Web Interface) doesn't have support for Flux-related architectures. We do expect to see a larger jump in image diversity via FLUX, as it supports significantly large prompt lengths; in our system, larger prompts result in more variance.

\subsubsection{Configuration Scale can systematically improves Diversity}

\begin{figure} [H]
    \centering
    \includegraphics[width=1\linewidth]{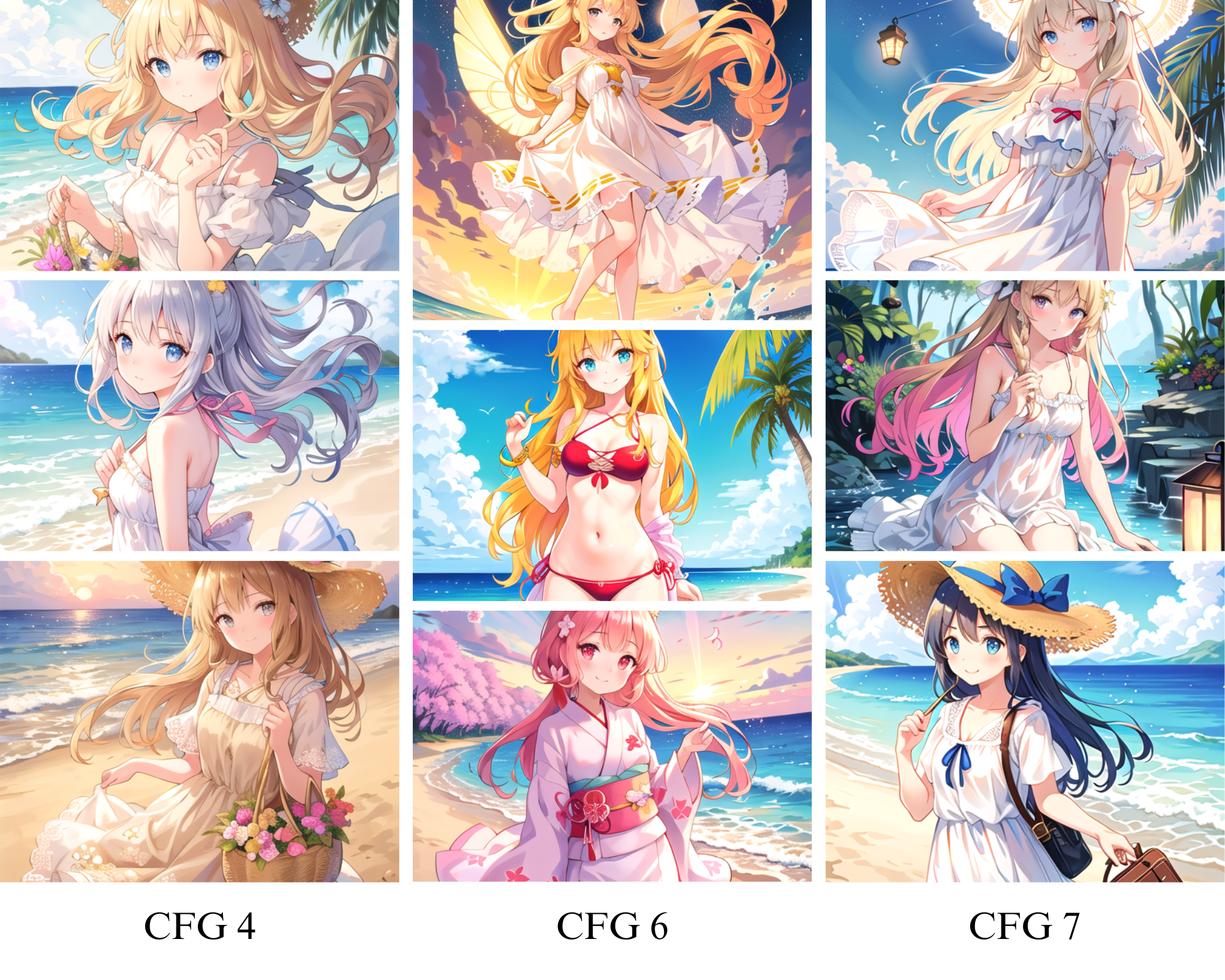}
    \caption{Examples of images generated with MANTA with various CFG values. SD 1.5 is used as the base model in this case. Prompt: \textit{a girl playing on the beach}}
    \label{fig:cfg-comparison}
\end{figure}

Traditionally, checkpoint users have used the \textit{CFG scale} argument in order to control how closely a checkpoint model aligns with a given prompt, thus reducing image diversity. 

However, with the prompt enhancement through concept mapping, we witness trends of image diversity \textit{increasing} as the CFG scale increases. This is because we provide prompt-induced variance, which, when followed closely, generates variance in the images as much as requested by prompting. 

Conversely, images from lower CFG values tend to approximate the model's natural variance via generation. In other words, lower CFG images obtain the majority of the variance from the model  itself, while high CFG models closely follow prompt variance. 

Ultimately, the choice of how much variance should one relagate to the model is left to the user. We do want to close this discussion by pointing out that prompt-based variance appears to be more than model induced variance, as compared between CFG 7 and CFG 4 images \ref{fig:cfg-comparison}:
\begin{itemize}
    \item CFG 7 images constantly vary their background scene, whereas CFG 4 consistently follows the theme of the beach
    \item CFG 7 images demonstrate more significant changes in the characteristics of the main subject (the girl)
    \item CFG 7 images also include handhold baskets, and more varying poses 

\end{itemize}

\section{Conclusion}

In this paper on MANTA, we showcase a system that seeks to provide users with more command over image diversity and quality while retaining alignment. To tailor our system towards the broader audience of AI artists, we focus on a GPU + RAM memory efficient retrieval augmented generation system that delivers reasonable performance. 

As evidence of it's capabilities, we witness MANTA gaining nearly a $50\%$ image diversity and quality win rate versus a normal image generation Stable Diffusion 1.5 model, and Stylus. The improvements in image diversity and quality do come with a minor cost to alignment, with the system marginally outpacing a base Stable Diffusion model ($51\%$) and coming up short with Stylus ($43\%$).

This iteration contains $40\%$ improvement in LLM token API usage and optimizations to drive down reliance over large amounts of data, paving the way for an eventual system that can be powered by completely open source LLM models without huge requirements for a large context length. 

This is done through a computationally efficient concept mapping structure, which systematically inserts configurable amounts variance into an image, which, as seen by the CFG variable can be tuned by the user to their usage. This system also provides an early example of a triplet loss-like mechanism for the document retrieval problem. 

\section{Discussion}

\subsection{Consumer Optimizations}

This paper has placed a strong emphasis on consumer feasibility, and in that, the system has also made concessions in the process that can lead to marginal performance gains if turned off. Specifically, in our implementation, all vector embeddings are quantized to the INT8 format, and sparse embeddings are emphasized to prevent extreme RAM usage.

Additionally, to further minimize token usage, we mapped each checkpoint or adapter to a single prompt. If multiple prompts were used, or multiple prompts were stacked in a single document, there are chances the results might be more relevant. 

\subsection{Next Steps}

\subsubsection{Performance Improvements}

The large areas of growth reside within the area of improved alignment. As illustrated by the Stylus paper in a CLIP vs FID Pareto curve,  there exists a tradeoff between diversity and alignment -- highly diverse images likely have lower prompt-image alignment, which may be consequential in narrow use cases. Across both of the retrieval based methods we see so far (alignment 

We attempted to experiment with using VLMs as a "postprocessor", iterating on taking in images, running img2img, and then modifying the CFG value to improve said alignment. However, vision LLM based control wasn't effective due to two primary issues - subjectivity and lora controllability. Our postprocessor algorithm would prompt the VLM to come up with a rubric-based response to score where to improve, and then correct a set of attention weights (which would either bring something into relative focus, or relax its significance in the prompt) by multiplying it by a scaling factor. Unfortunately, the VLM's subjectivity led it to critically receive many concepts in the concept mapping framework as not showing up, which led to extremely large concept weights that eventually eroded visual fidelity. 

Another point of interest would be exploring various LoRA recommendation policies. Currently, the concept mapping framework boils down the name, detail, and styles of each concept, which are then queried against the database of adapters. While this leads to aggregate, concept-level control, it would be worthwhile to see if lower level control such as querying based on styles and details can provide even better improvements in image diversity.

It would also be of interest to see if adding additional human-in-the-loop mechanisms could provide lower level control without too much additional human input. As the prompt and generation are closely powered by an LLM, humans could ask the LLM in a standard chat interface to "\textit{Generate photos of XYZ}", and then iteratively refine them through further img2img results. We also believe that while alignment would be a lofty goal to pursue, human intervention in small, iterative amounts would be a functionally adaquate solution for improving alignment results.

\subsubsection{Future Development}

We have intentionally set up a budget efficient adapter system in order to ensure that other models can effectively be switched in. In future work, we hope to conduct evaluations on various standard open source LLMs being used in place, and testing a 'completely open source' workflow. 

Envisioning demand for further control of generation for niche tasks, we also hope to integrate more autonomy into finetuning if no information is provided, likely through determining criterion in which it would be optimal to create additional LoRAs to systematically ensure a concept is "well defined" within the scope of a generation. For example, if a sports game is sought to be replicated via AI images, such an algorithm might include LoRAs for essential parts like the tennis net, or the various court structures.

Finally, we look forward to a path where the system can cost-effectively become multimodal, being able to factor images and styles as validation for checkpoints previously used. This would enable the system to become closed-loop, as images can be generated, and then assigned a rating that would improve retrieval results.

\subsection{Reproducibility}
We include steps here to reproduce key results that we have cited in our paper. As further attempts at reproduction are performed, we hope to improve this section as per their feedback. 

In order to reproduce evaluation over the main results, one can visit the link here: \url{https://drive.google.com/file/d/1NhCmI05nYCNq4luWNvRddfKz3QNCwtIS/view?usp=sharing}. This link will contain a zip file to the generation run, set for $500$ different prompts on COCO 2014. Please download the zip file, and run the automated evaluation script in our codebase \ref{section:code} with the diversity, alignment, and quality parameters for the respective tasks.

\subsection{Code}
\label{section:code}
A publicly available repository for the Python code will be found on GitHub at the following link: \url{https://github.com/AnshKetchum/stylus2}. We are still in the process of removing developer-centric debugging print statements, and other artifacts to make the codebase more readable and presentable. 

\section*{Acknowledgments}
We thank Michael Luo for his gracious time in providing constructive feedback, and perspective as the original author of the Stylus 1.0 paper. We also thank Saarthak Kapse from Stonybrook University for his kind guidance and domain-specific suggestions on additional ablations. 
\end{multicols}

\newpage
\appendix
\section{Appendix}
\subsection{Use cases}
\textbf{AI Art.} We discuss some of the examples of how this system can be used to creatively come up with high quality, diverse images that may serve as starting points for further refinement. Our experiments do show case a clear usecase for MANTA in the AI art area. 

We find that MANTA can generate stylistically diverse images that also feature characters in various poses and backgrounds, the results of which have been previously discussed. In particular, we find that the system is especially effective at creating images within the anime area, primarily due to the large amount of training data hosted on the platform in the niche. 

\begin{figure}[H]
    \centering
    \includegraphics[width=1\linewidth]{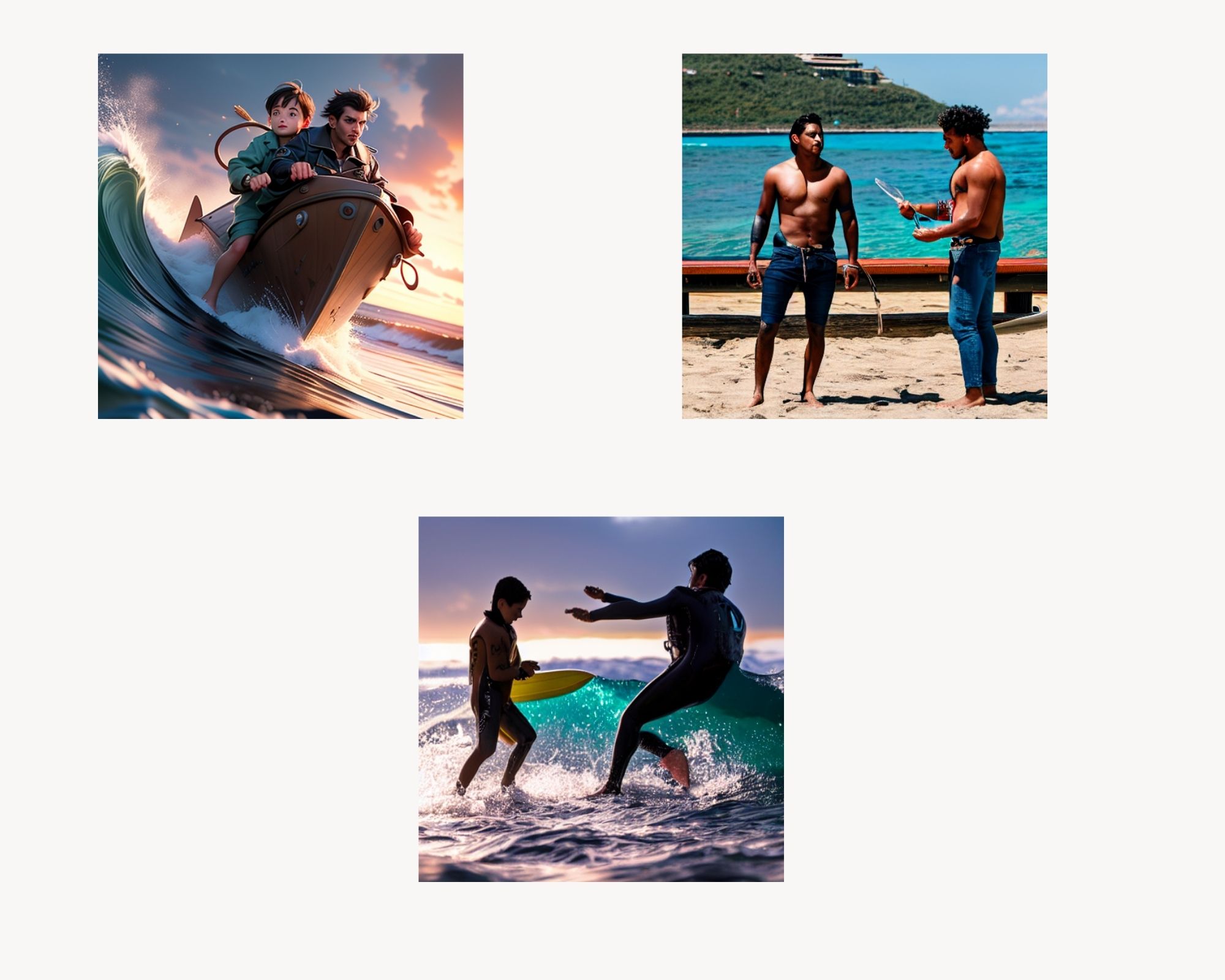}
    \caption{MANTA output for the prompt - "Teaching a boy how to surf". There is high variance across style and character action. }
    \label{fig:art-usecase}
\end{figure}

Within this area of prompting, including closely associated concepts, even vague prompts resulted in high quality images. For example, a simple prompt such as "a man teaching a boy how to surf" came up with the variety shown \ref{fig:art-usecase}. 
However, we find that while these images may be high in quality, these image leave enough room for additional improvement through inpainting, upscaling, high-resolution (hires), as well as touch ups such as face restoration. 

For example, in the human case we found that parts of human body details in the image may be malformed. These are commonly the hands, fingers, nose, lips, or any parts that would seem like "small details" in an image but be immediately noticeable if incorrectly portrayed.

\begin{figure}[H]
    \centering
    \includegraphics[width=0.5\linewidth]{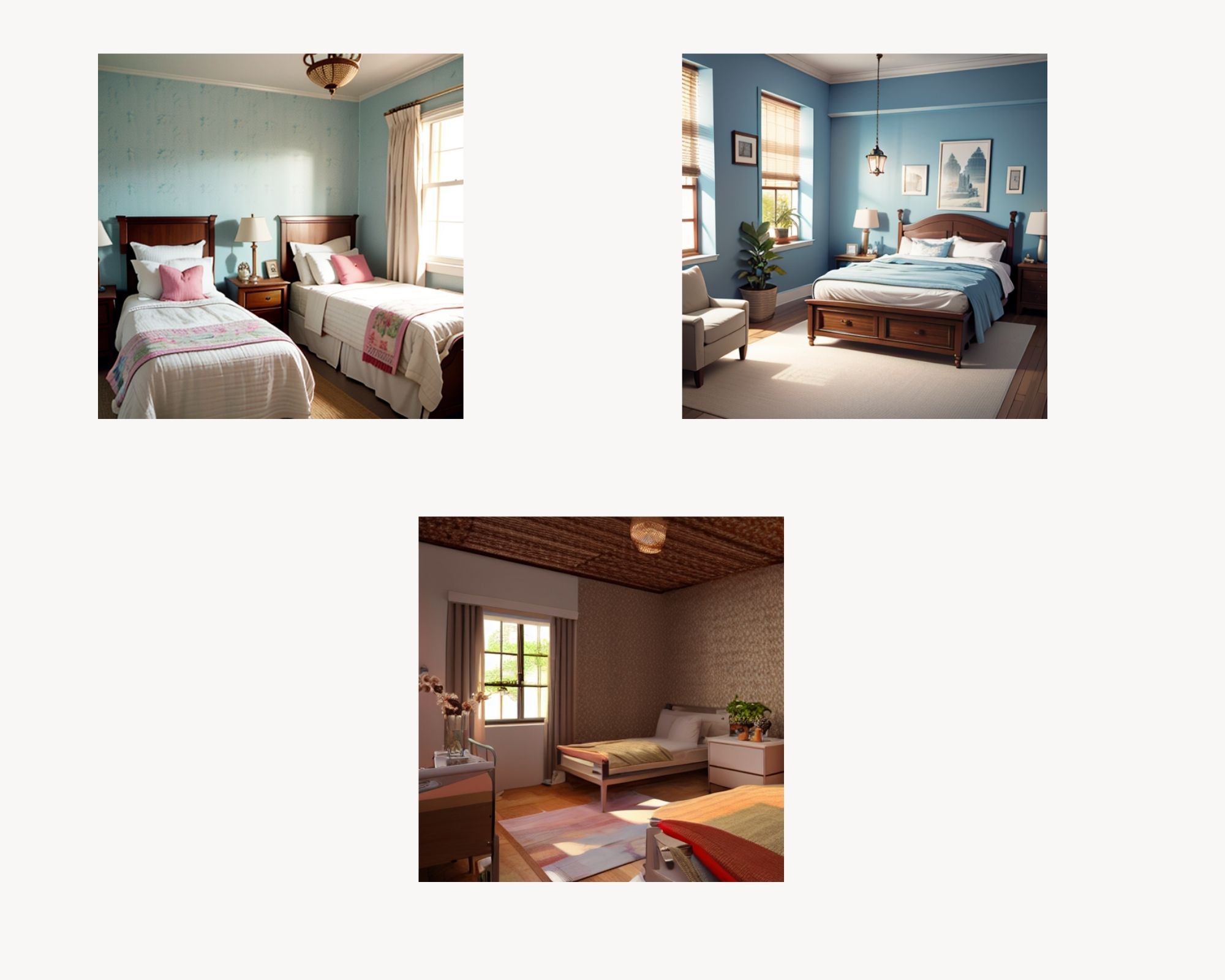}
    \caption{MANTA output for a synthetic generation usecase involving furniture. Prompt: A bedroom with twin beds and linen. }
    \label{fig:synthetic-data}
\end{figure}

\textbf{Synthetic Data Generation.} We do see a strong use case of MANTA as a synthetic "data factory", which leverages checkpoints and LoRAs to create highly diverse images. We present some realistic example that may pass off with some minimal tolerance as a potential training data sample \ref{fig:synthetic-data}.

We also present a simple procedure for any company interested in expanding their dataset with synthetic images using MANTA. Typically, AI models require images at a large scale - between 100K to 1M image on average - for state of the art performance, and we provide a procedure of how Stylus can be used to achieve these gains. 

\textbf{Synthetic Data Expansion Algorithm}

\begin{enumerate}
    \item Find a dataset $D$ resembling the type and niche of your images. Common examples include COCO 2014 \cite{lin2015microsoftcococommonobjects}, Pascal VOC, Roboflow 100K \cite{ciaglia2022roboflow100richmultidomain}, etc. 

\textit{Note that these datasets must contain two requirements to be a viable option - images and a prompt "caption" for each image} 

    \item Pick a series of adapters $A$ and a set of checkpoints $C$ to be able to create data. 

    \item Run Stylus on $D$, provided the adapter set $A$, and checkpoint set $C$. Obtain output coarse dataset of images $D'$

    \item Run finetuning to ensure that images in $D'$ are reasonable additions to the data distribution of the previous dataset, $D$. 

A simple example of a \textit{postprocessing algorithm} could be as follows:

\begin{spverbatim}
    def postprocess(D, D'):
        set FID real images to D
       for each generated_image_set in D':
          compute FID with the new image_set per caption

          if FID < threshold:
            add to D
    
\end{spverbatim}
\end{enumerate}

\subsection{Failure Modes}
In this section, we discuss various failure modes discovered while developing MANTA. 

\subsubsection{Concept Overload.} 
\begin{figure}[H]

    \includegraphics[width=1\linewidth]{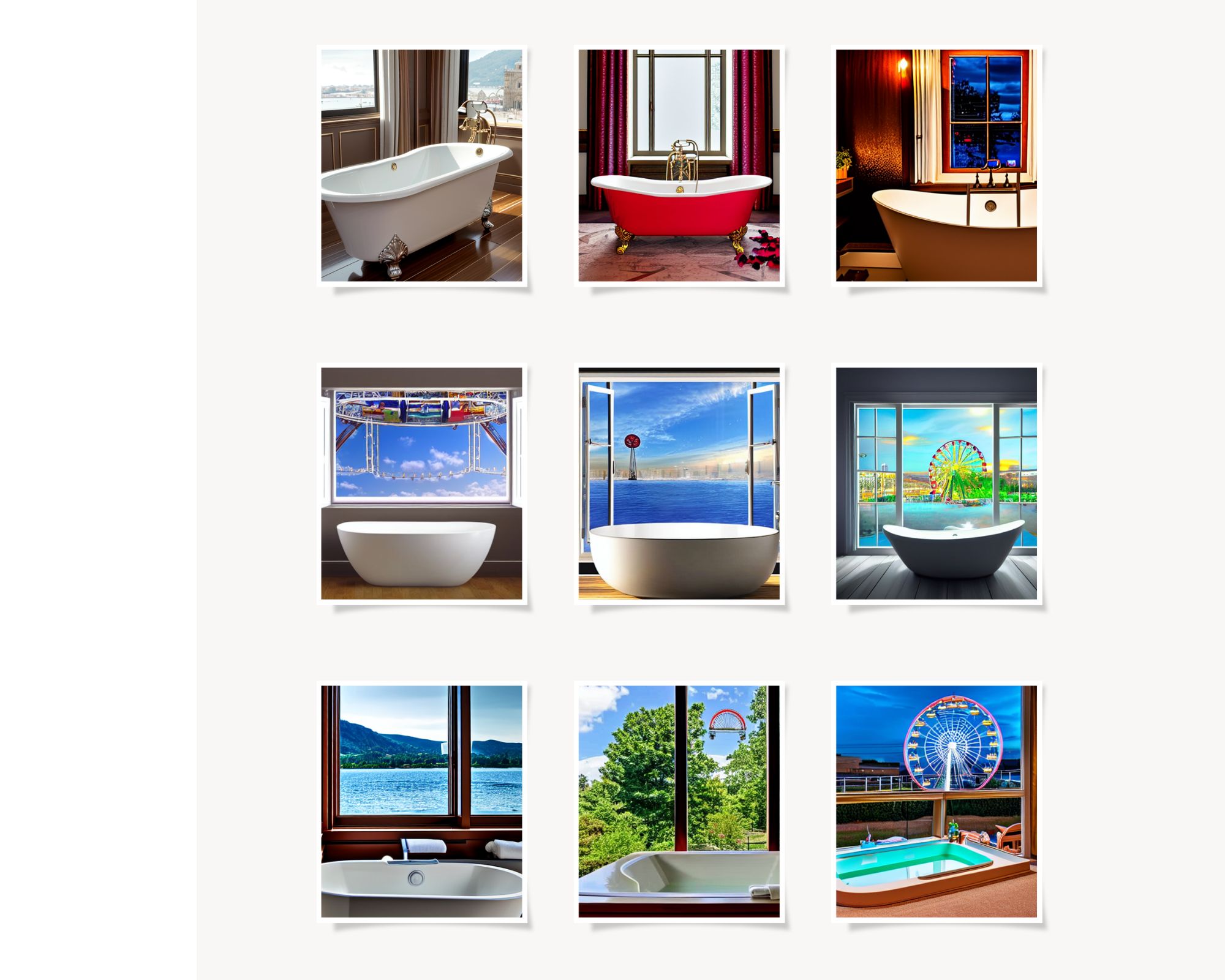}
    \caption{Example of images for the prompt - \textit{A bathtub sits next to a ferris wheel}, from MANTA, Stylus, and base Stable Diffusion, separated by rows. The prompt demonstrates an example of \textit{concept overload}, where one concept is given overwhelming priority with a second concept marginalized or omitted. }
    \label{fig:concept-overload}
\end{figure}

In this situation, MANTA would over-focus on one concept, ignoring or omitting other concepts that would have been relevant. Oftentimes, we found this would happen when it would be challenging to pick a leading \textit{main} concept. This led to a concept perpetually being subdued to the background or a corner of the image. In the case of latest iteration of MANTA, it would often ignore the secondary concept entirely, resulting in higher fidelity images at a heavy alignment penalty for ignoring a concept \ref{fig:concept-overload}. The other algorithms attempted to superimpose the two concepts, leading to low quality images where the second object wasn't often created with the high quality of the first.

\subsubsection{Concept Relationship Misunderstanding.} 
\begin{figure} [H]
    \centering
    \includegraphics[width=0.5\linewidth]{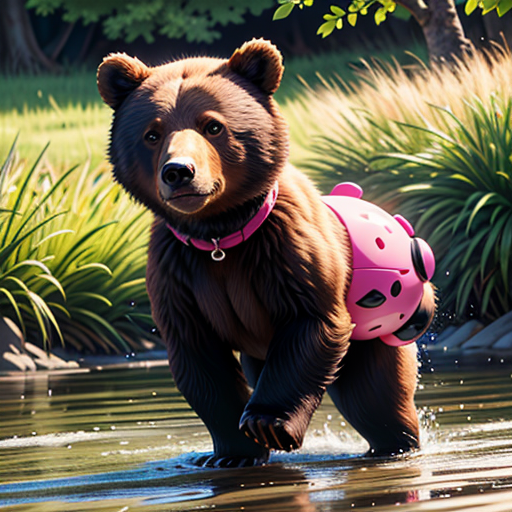}
    \caption{MANTA output for \textit{A bear carries a pink ball by the river side}. While the image does demonstrate diversity, the concept "ball" (the pink object wrapped around the bear) loses its meaning. }
    \label{fig:concept-misalignment}
\end{figure}

Similar to the previous concept, this issue occurs when two concepts intersect. If the model doesn't seem to have experience understanding how to relate the two objects, it results in naive merger or low diversity output. An example of a failing prompt for this case is \textit{A bear carries a pink ball by the river side} \ref{fig:concept-misalignment}. While the output can be considered to be novel and artistically diverse, the diversity isn't sensibly generated - the interactions of the concepts aren't in the realm of possibility one would expect. 

\subsubsection{Adapter Gating.} In this situation, a lack of highly relevant adapters causes the retrieval adapters to produce results with weaker retrieval scores. The Stylus algorithm for LoRA selection can be described by the following pseudocode: 

\begin{spverbatim}
def query_loras(prompt,
support_embeddings, k, init_thresh):
    load adapters from database D 
    returned_loras = {}
    threshold = init_thresh

    while we do not have k returned_loras:
        returned_loras = retrieve k adapters
        returned_loras = filter out adapters < threshold

        if not enough adapters:
            threshold = 0.95 * threshold

    return returned_loras
\end{spverbatim}

However, we experienced edge cases while testing with Stylus Docs, where the adapters returned would often be zero due to the lack of relevant adapters. This would cause cases where the checkpoint would often be limited by it's training knowledge of the concepts for task image diversity. 

\subsubsection{Rogue adapter / checkpoints.} This problem is exacerbated by the fact that we have the capability to sample from two sets $C$ and $A$, rather than one set of adapters $A$. This problem still significantly perturbs our system, as there is no algorithm employed for ensuring high quality adapters. We uphold previous work implementations by providing manual guardrails through adapter/checkpoint blacklists and word based filters. 

\subsection{Prompts}

\subsubsection{Detail Enhancement}
\label{prompt:detail-enhancement}
\begin{spverbatim}

    """You are helping a candidate create prompts for an image generation model.

    Below is an input containing some sort of concept. The concept could be an entity, idea, noun, anything of that sort. Your job is to create a bulleted list of details that further make the idea more tangible. Think of things that the concept is composed of, and then provide me a list of {n} extremely specific details for that concept. 

    Here are some examples of extremely detailed responses: 

    - Concept: anime 

    Response: anime girl, white hair, red dress, thigh highs, slim chest

    - Concept: samurai

    Response: samurai robot warrior,  large traditional straw hat, 

    Note: Make sure the details directly add to the current character rather than creating new ones. Here is a bad example of details that create additional characters:

    Create a comma separated list of {n} extremely specific details that provide extremely vivid clarity to the concept. Do not include any verbs at the moment.

    Concept: {concept}"""
\end{spverbatim}

Above is an example of a version of the detail refinement prompt that we fed to an LLM to generate a list of $n$ further details. Using this prompt, we can systematically increase or decrease specificity within the prompt.

\bibliographystyle{unsrt}  
\bibliography{references}

\end{document}